# SKETCH LESS FACE IMAGE RETRIEVAL: A NEW CHALLENGE

*Dawei Dai, Yutang Li, Liang Wang, Shiyu Fu, Shuyin Xia, Guoyin Wang*\*

College of Computer Science and Technology, Chongqing University of Posts and Telecommunications
Chongqing 400065, China
wanggy@cqupt.edu.cn, dw_dai@163.com

## ABSTRACT

In some specific scenarios, face sketch was used to identify a person. However, drawing a complete face sketch often needs skills and takes time, which hinder its widespread applicability in the practice. In this study, we proposed a new task named **sketch less face image retrieval** (SLFIR), in which the retrieval was carried out at each stroke and aim to retrieve the target face photo using a partial sketch with as few strokes as possible (**see Fig.1**). Firstly, we developed a method to generate the data of sketch with drawing process, and opened such dataset; Secondly, we proposed a two-stage method as the baseline for SLFIR that (1) A triplet network, was first adopt to learn the joint embedding space shared between the complete sketch and its target face photo; (2) Regarding the sketch drawing episode as a sequence, we designed a LSTM module to optimize the representation of the incomplete face sketch. Experiments indicate that the new framework can finish the retrieval using a partial or pool drawing sketch. (https://github.com/ddw2AIGROUP2CQUPT/SLFIR)

*Index Terms*— Sketch-Based Image Retrieval; Face Sketch; Partial Sketch; Sequence Learning;

## 1. INTRODUCTION

Face is the most common biometric used for the identification of a person. Because of the requirements of some practical application, cross-domain face retrieval based on sketch has aroused widespread interest. For example, the sketch painting of visual description given by the onlooker are commonly used by the law enforcement agencies to identify the suspect; Face sketch also plays an important role in the teaching of art majors. In addition, considering the rapid proliferation of various electronic touch screen devices, more convenient hand-painted input conditions have been provided for the majority of users, Sketch-based face retrieval will also have broad application prospects in daily work and life.

Consequently, sketch-based image retrieval (FG-SBIR) address the problem of retrieving a particular photo for a given query sketch. For such problem, some methods carries out the retrieval by designing the efficient feature description operators with invariance for cross-domain images[1-7]; some methods are to convert sketches into images by designing transfer model[8-13] and then carry out the retrieval task;

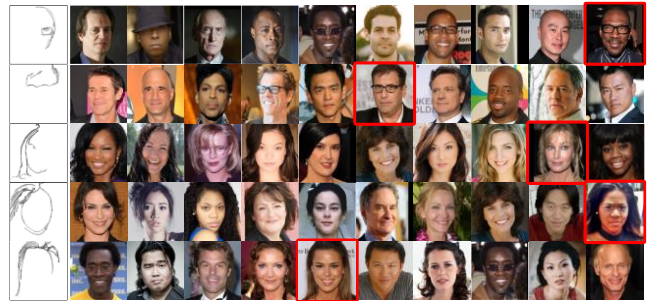

**Fig.1.** Demonstration of some instances of our proposed **SLFIR**. Target face photo can be correctly retrieved in Top-10 list when few strokes are used.

some methods aim to map the sketch and photo into the same embedding space, in which we can directly calculate the similarity for sketch and image using their embedding vectors [14-19].

However, drawing a face sketch often takes time and requires skills. Considering the problems, we proposed a new framework named SLFIR that the retrieval begins after each stroke and aim to retrieve target face photo using as few strokes as possible (**as shown in Fig.1**). SLFIR framework can provide some form of interaction or feedback for human, which may have greater application potential in many fields, such as sketch teaching, criminal investigation, entertainment, and so on.

Since, training a model for SLFIR requires additional data of face sketch with drawing process; As a result, first, we presented a method to simulate the face sketch drawing process based on the complete sketch (**as shown in Fig.4**), and open two datasets. Second, we proposed a two-stage model as the baseline method for SLFIR problem, in which the first stage learn the joint embedding space for the complete sketch and photo, and the second stage optimize the embeddings of the partial sketch. Experiments show that our proposed SLFIR can achieve the retrieval using the partial or poor drawing sketch. Our contributions can be summarized as follows. (a) We proposed a new framework for sketch-based face image retrieval to break the barriers that hinder its widespread applicability; (b) We presented a baseline method for the SLFIR problem, and can perform better at the early retrieval than that of classical method.

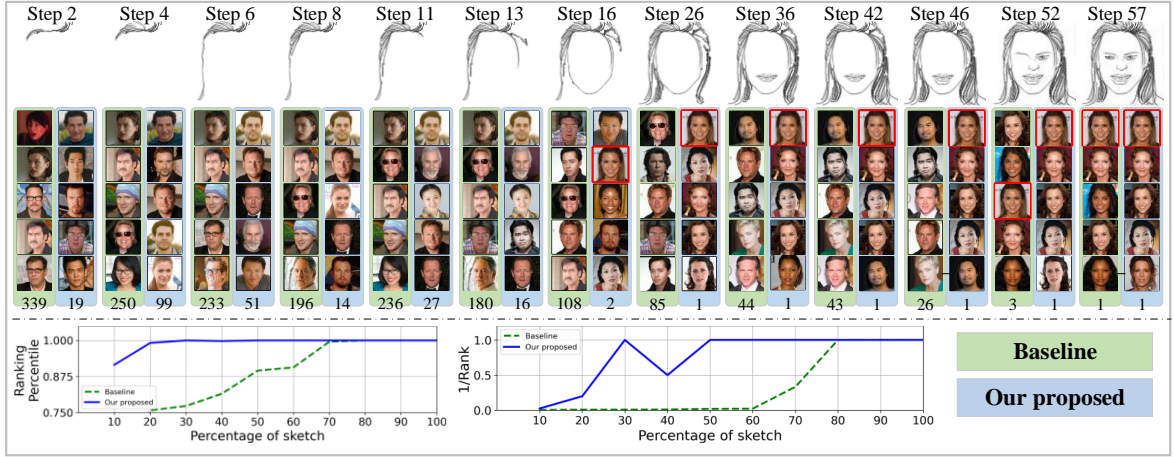

**Fig. 2.** Illustration of SLFIR framework. Our proposed and a baseline for FG-SBIR (B1)[20] that trained on only the completed sketches. Our proposed method can retrieve the target photo using fewer strokes than that of B1. The number at the bottom denotes the paired (true match) photo's rank at every stage.

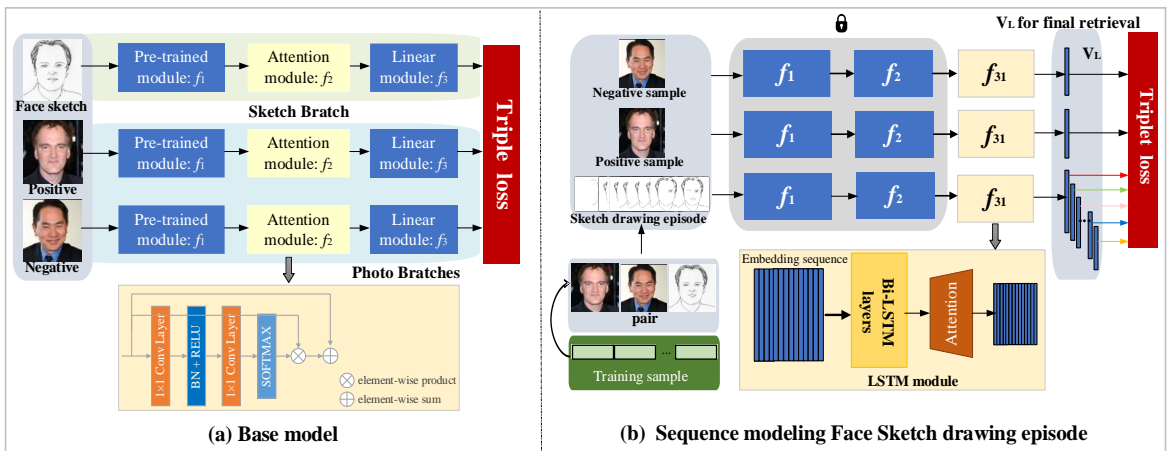

**Fig. 3.** Overview of our approach. (a) Base model: a classical FG-SBIR framework; (b) Our proposed for SLFIR. The locks signify that the weights are fixed during learning.

## 2. METHODOLOGY

For the SLFIR, we presented a baseline method to learn the efficient representations for the partial face sketches and face photos. We first require such a dataset that includes face photos and their corresponding sketch drawing process. In fact, it is difficult for us to obtain a large amount of such data. In this section, we first proposed a method to generate face sketch drawing episode based on the complete sketch; and then, we proposed a two-stage method to address such SLFIR problem.

### 2.1 Generating the sketch drawing episode

We proposed a method to generate the data of sketch drawing process. First, we adopt a pre-trained CNN model[21] to extract structural lines for each complete sketch. Second, we use canny operator[22] to detect edge lines, and store the edge lines that exceed the threshold of a certain number of pixels. At last, (1) We repaint the edge sequence incrementally on the blackboard based on some agreed strategies and save it until all the edge sequences are drawn; (2) Since, the width of edge line extracted by Canny was only a single pixel, and thus Dilation operation was used to combine these edges into thick line; (3) In order to further simulate the drawing style of a sketch rather than an edge image, we do the dot multiplication between the repainting edges and the original region of the complete sketch. One instance was shown in **Fig.4**.

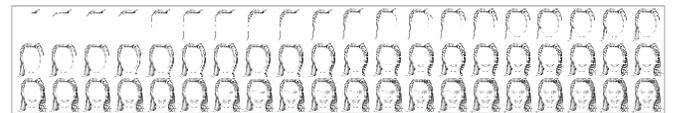

**Fig. 4.** Illustration of a face sketch drawing process generated by our proposed.

### 2.2 A Base Framework for SLFIR

#### 2.2.1 Overview

We consider the SLFIR framework as a sequence optimization problem and propose a method of combining CNN and LSTM to deal with the sketch drawing episode, in which CNN module was used to extract the feature for input image, and LSTM module was used to optimize the representation of the incomplete face sketch. We aim to retrieve the target face photo at the earliest stroke possible (**see Fig.2**). An overview of our proposed model is shown in **Fig.3**. We first train a state-of-the-art FG-SBIR model[23] based on the complete face sketches and photos. Next, we learn the sketch branch using a Bi-LSTM module and triplet loss for the incomplete face sketch sequence. For a query of sketch $s$, we obtain its embedding vector (d-dimensional feature vector) using our proposed (**see Fig. 3(b)**) and obtain the top-$k$ list based on the pairwise distance metric.

*2.2.2 Base Model*
First, a neural model was designed to learn the joint embedding space for complete face sketch and its target photo. To verify the effectiveness of our framework, we use a classical triplet network, containing three CNN branches with shared weights corresponding to the input face images, including a positive photo, a query sketch (complete one), and a negative photo, as shown in **Fig.3(a)**. The model can be divided into three parts. The first part is a pre-trained CNN model $f_1$, which was used to extract the features of the input images. Here, InceptionNet[24] trained on ImageNet performed the function $f_1$ (see Eq. (1)), where $x$ and $B$ indicate the input image and its corresponding feature maps, respectively. The second part $f_2$ allows learning the embedding vector for the complete face sketch and its target photo, as shown in Eq. (2); here, we use soft spatial attention[25]. The third part $f_3$ maps the high-dimensional vectors ($V_H$) to low-dimensional vectors ($V_L$), using a simple linear mapping ($A$) in Eq. (3). We applied a triplet loss on each pair (complete sketch, target face photo, non-target photo). The terms $v_{SL}$ and $v_p$ ($v_n$) in Eq. (4) indicate the low-dimensional vector of the input pair obtained using Eq. (3).

$$B = f_1(x) \quad (1)$$

$$V_H = f_2(B) = Global\_pooling(B + B \cdot f_{att}(B)) \quad (2)$$

$$V_L = AV_H \quad (3)$$

$$J_1(\theta) = \max(\sum_0^n d(v_{SL}, v_p) - d(v_{SL}, v_n) + \alpha, 0) \quad (4)$$

*2.2.3 Bi-LSTM Modeling Face Sketch Drawing Episode*
We designed another learnable module to optimize the representations for the incomplete sketches. In our work, dealing with the drawing episode was considered as a sequence optimization problem. As shown in **Fig.3(b)**, we designed a Bi-LSTM module to learn the temporal correlation in the drawing episode and optimize the representation. First, we train the base model to obtain the feature vector $V_H$ (high-dimensional, see Eq. (2)) for all incomplete sketches ($S = \{s_i\}$, $i=1,2,...,T$) in a drawing episode; second, we consider $\{V_H[i]\}$ as a temporal sequence, and adopt a Bi-LSTM module (including several Bi-LSTM layers) and a triplet loss function (see Eq. (4)) to learn the embedding space of the incomplete sketch as close to its target photo. Here, the term $V_{SL}$ indicates the low-dimensional vector of the incomplete sketch in a drawing sketch obtained Eq. (5).

$$V_{SL} = f_{lstm}(\{V_H[i]\}), i = 0,1,2,...,T \quad (5)$$

**2.3 Evaluation Metric**
Regarding FG-SBIR, we prioritize the target photo appearing at the top of the list. As mentioned in[20], m@A (the ranking percentile) and m@B (1/rank versus percentage of sketch) were used to capture the retrieval performance for the incomplete sketches. In order to better reflect the early retrieval performance, we developed the weighted m@A and m@B (as shown in Eq. (7)) that the earlier the sketch, the greater the weight of its performance was. In Eq. (6), terms of $p_i$ and $p_n$ indicate the number of strokes of current sketch and complete sketch. In Eq. (7), term $m$ indicates that test data contains $m$ photos, term $n$ indicates that one sketch drawing episode correspond $n$ incomplete sketches; term $rank_{ji}$ indicates the rank of target photo based on the $i^{th}$ sketch of $j^{th}$ photo. A higher value of all metrics indicates a better performance during the early sketch retrieval.

$$w_i = e^{-\frac{p_i}{p_n}} \quad (6)$$

$$w@mB = \frac{\sum_j^m \sum_i^n w_{ji} \frac{1}{rank_{ji}}}{m \cdot n}, \quad w@mA = \frac{\sum_j^m \sum_i^n w_{ji}(1 + \frac{1 - rank_{ji}}{m-1})}{m \cdot n} \quad (7)$$

## 3. EXPERIMENTS

**Dataset.** We chose two sub-datasets (named D1 and D2, drawn by different painters) from FS2K[26] as the base dataset. Since, original FS2K contains only complete face sketches; to address the SLFIR, we generate the data of sketch with drawing process based our proposed. Here, each sketch drawing episode contains a sequence of 70 incomplete sketches. Therefore, D1 contains 107,030 sketches and 1,529 photos, of which 75,530 sketches and 1,079 photos are used for training and the rest for testing. D2 contains 33,390 sketches and 477 photos, of which 23,380 sketches and 334 photos are used for training, and the rest for testing.

**Implementation details.** We implement our model on a 40GB NVIDIA A100. Training process is divided into two stages. In stage1, to train the base model, we used the triplet loss with margin of 0.3 and Adm optimizer: for Inception-V3 part, the convolutional layer parameters are fixed for all non-Inception blocks, while the Inception blocks are updated with a learning rate of 5e-4, and attention module and fully connected layer initialized using Kaiming normal[27] are updated with a learning rate of 5e-3. The mini-batch size of stage1 is 64 (D2 is 32). In stage2, we fixed the Inception-V3 part and attention module of stage1, and then we also used a triplet loss with margin of 0.3 and Adam optimizer with a learning rate of 5e-4; Then, we input sketch sequence images with mini-batch size of 32 and T=70 steps into the network, and train a total of 300 epochs. The channel of the extracted feature map for Inception-V3 in stage1 was 2048, and the hidden state of the stage2 through the Bi-LSTM layer was 1024.

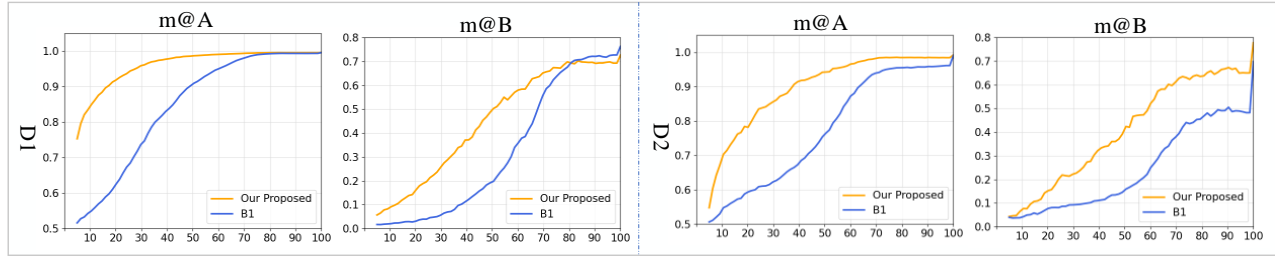

**Fig. 5.** Performance of early retrieval. Instead of showing the complete sketch, we visualize it using the percentage of sketch. A higher value indicates a better early retrieval performance.

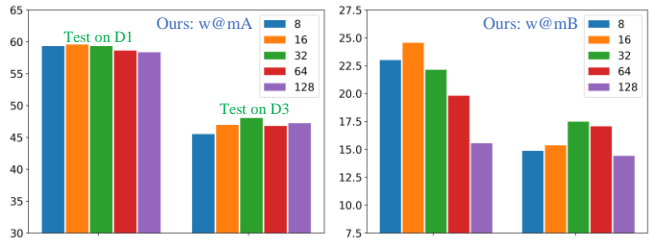

**Fig. 6.** Performance of ours and B1 on the Testdata of D1 (Left part) and D3 (Right part) respectively. In each subfig, 5 colors represent feature embedding with different dimensions (8, 16, 32, 64, 128).

### 3.1 Performance Analysis

The performance during the drawing process of our proposed two-stage model on the problem of SLFIR is shown in **Fig.5** against the baseline method (B1: The framework is the same with our base model that trained only with a triplet loss using complete sketch, shown in **Fig.3(a)**). We can note that (1) In SLFIR framework, we can expect to obtain the better performance using fewer strokes as possible through designing the suitable models, so that users can retrieve the target photo efficiently; (2) Our proposed two-stage method performs much better at early retrieval than that of B1; Compared to the B1, we design a second-stage learnable module to optimize the representations of the incomplete sketches that obtained from first stage without confusing the base model.

**Table 1.** Comparative results with varying feature-embedding on **D1** and **D2**

|    | Dim | B1 | | Ours | |
|----|-----|------|------|------|------|
|    |     | $w$@mA | $w$@mB | $w$@mA | $w$@mB |
| D1 | 16  | 50.40 | 15.83 | **59.57** | **24.56** |
|    | 32  | 49.86 | 17.13 | **59.33** | **22.13** |
|    | 64  | 48.73 | 15.08 | **58.63** | **19.81** |
|    | 128 | 52.97 | **17.70** | **58.34** | 15.54 |
| D2 | 16  | 46.00 | 12.47 | **54.85** | **22.22** |
|    | 32  | 44.19 | 15.43 | **53.43** | **21.36** |
|    | 64  | 44.14 | 13.47 | **53.13** | **18.73** |
|    | 128 | 47.56 | **15.40** | **51.60** | 15.30 |

All the quantitative results on Testdata are shown in Table.1, we observe that our proposed performs significantly better than that of the B1 at each dimension. We also noticed a phenomenon that the metric shows certain fluctuations for such problem. The reason can be noted that (1) The dimension of feature-embedding has different effects on sketches at different percentage, for example, high dimensional feature-embedding may play a positive role for complete sketches, but not necessarily for the incomplete sketches; (2) Not all strokes can contain positive information, some of strokes degrade the performance.

### 3.2 Further Analysis in the Practice

In this section, we aim to verify the performance of our proposed in practice. We invited 50 students (art major) to submit 100 sketch drawing episodes (named "D3"). As shown in **Fig. 6**, comparing with D1, the performance of both our proposed and B1 methods on D3 decreased, we consider that the diversity of painting process at early stage can be noted as one of the main reasons. Table.2 shows the comparative results of our proposed and B1, we can note that our proposed performs significantly better at easily retrieval than that of B1 in the practice.

**Table 2.** Comparative results with varying feature-embedding on **D3**

| Dim | B1 | | Ours | |
|-----|------|------|------|------|
|     | $w$@mA | $w$@mB | $w$@mA | $w$@mB |
| 8   | 41.30 | 10.85 | **45.52** | **14.87** |
| 16  | 41.90 | 11.62 | **46.95** | **15.36** |
| 32  | 42.70 | 13.68 | **48.05** | **17.49** |
| 64  | 41.32 | 11.33 | **46.79** | **17.06** |
| 128 | 42.75 | 12.93 | **47.22** | **14.42** |

### 4. CONCLUSION AND FUTURE WORK

Drawing a face sketch often takes more time and requires strong skills, which hinder its widespread application in the practice. We considered that interaction between human and machine can be able to improve the efficiency of retrieval. Inspired of this, we proposed a new framework named SLFIR performing retrieval after each stroke, in which human can obtain the feedback from retrieval and make an adjustment during the drawing process. For the SLFIR, we open three dataset of sketch drawing episodes, and then proposed a two-stage model. However, SLFIR problem face some major challenges. For example, (1) how to address the diversity of sketches at early stage; (2) how to narrow the cross-domain gap between the early incomplete sketch and its target face photo.